\documentclass[conference,compsoc]{IEEEtran}
\ifCLASSOPTIONcompsoc
  \usepackage[nocompress]{cite}
\else
  \usepackage{cite}
\fi
\ifCLASSINFOpdf
  \usepackage[pdftex]{graphicx}
\else
  \usepackage[dvips]{graphicx}
\fi
\usepackage{amssymb}
\usepackage{amsmath}

\usepackage[linesnumbered,vlined,ruled]{algorithm2e}

\begin{document}
%
\title{Real-Time Deep Learning Method for Abandoned Luggage Detection in Video}

\author{\IEEEauthorblockN{Sorina Smeureanu\IEEEauthorrefmark{1}\IEEEauthorrefmark{3},
Radu Tudor Ionescu\IEEEauthorrefmark{1}\IEEEauthorrefmark{3}}
\IEEEauthorblockA{\IEEEauthorrefmark{1}University of Bucharest, 14 Academiei, Bucharest, Romania}
\IEEEauthorblockA{\IEEEauthorrefmark{3}SecurifAI, 24 Mircea Vod\u{a}, Bucharest, Romania\\
E-mails: sorinasmeureanu@gmail.com, raducu.ionescu@gmail.com}}


%


\maketitle

\begin{abstract}
Recent terrorist attacks in major cities around the world have brought many casualties among innocent citizens. One potential threat is represented by abandoned luggage items (that could contain bombs or biological warfare) in public areas. In this paper, we describe an approach for real-time automatic detection of abandoned luggage in video captured by surveillance cameras. The approach is comprised of two stages: $(i)$ static object detection based on background subtraction and motion estimation and $(ii)$ abandoned luggage recognition based on a cascade of convolutional neural networks (CNN). To train our neural networks we provide two types of examples: images collected from the Internet and realistic examples generated by imposing various suitcases and bags over the scene's background. We present empirical results demonstrating that our approach yields better performance than a strong CNN baseline method.
\end{abstract}


%
\IEEEpeerreviewmaketitle

\section{Introduction}
\vspace*{-0.2cm}

Recent terrorist attacks in major cities around the world have brought many casualties among innocent citizens\footnote{{https://en.wikipedia.org/wiki/Boston\_Marathon\_bombing}}. In this context, we need systems able to provide real-time alerts about potential threats, so that people can enjoy life in their own cities without being afraid of terrorist attacks. In this paper, we deal with the problem of abandoned luggage detection. When an object, usually a suitcase or a bag, is left unattended in a public area, e.g. airport terminal or train station, it represents a security threat because the abandoned object may contain dangerous items such as explosives or biological warfare. For this reason, the abandoned object must be removed immediately from the public area by authorized personnel. To this end, we propose a system that is able to automatically detect abandoned luggage items in real-time by analyzing the video from surveillance cameras. Our approach is comprised of two stages. In the first stage, we detect static objects based on background subtraction and motion estimation, as most related works~\cite{Bhargava-AVSS-2007,Liao-AVSS-2008,Porikli-JASP-2008,Tian-VS-2008,Bhargava-MVA-2009,Wen-ICIP-2009,Chang-JASP-2010,Forczmanski-ICCVG-2010,Kwak-OE-2010,Szwoch-IIMSS-2010,Tian-TSMC-2011,Lin-ICPR-2014,Lin-TIFS-2015,Dahi-CVIU-2017,Pham-ICISP-2017}. In the second stage, we apply a cascade of \emph{convolutional neural networks} (CNN) based on the GoogLeNet~\cite{Szegedy-CVPR-2015} architecture to distinguish between abandoned luggage items and other objects, e.g. persons standing still. To obtain robust CNN models, we provide two types of examples: images collected from the Internet and realistic examples generated by imposing various suitcases and bags over the scene's background. To our best knowledge, we are the first to train a cascade of convolutional neural networks for abandoned luggage recognition. 

The paper is organized as follows. Related work on abandoned luggage detection is presented in Section~\ref{sec_RelatedWork}. Our learning framework is described in Section~\ref{sec_Method}. The abandoned luggage detection experiments are presented in Section~\ref{sec_Experiments}. Finally, we draw our conclusions in Section~\ref{sec_Conclusion}.

\section{Related Work}
\label{sec_RelatedWork}
\vspace*{-0.2cm}

Most works for abandoned objects detection use background subtraction as a low-level preliminary step~\cite{Bhargava-AVSS-2007,Liao-AVSS-2008,Porikli-JASP-2008,Tian-VS-2008,Bhargava-MVA-2009,Wen-ICIP-2009,Chang-JASP-2010,Forczmanski-ICCVG-2010,Kwak-OE-2010,Szwoch-IIMSS-2010,Tian-TSMC-2011,Lin-ICPR-2014,Lin-TIFS-2015,Dahi-CVIU-2017,Pham-ICISP-2017} to detect foreground regions or objects, although Smith et al.~\cite{Smith-PETS-2006} start directly by tracking multiple objects in the scene using trans-dimensional Markov Chain Monte Carlo (MCMC). After background subtraction, some works aim to reduce false positive detections using object tracking and classification methods~\cite{Bhargava-AVSS-2007,Liao-AVSS-2008,Chang-JASP-2010,Forczmanski-ICCVG-2010,Kwak-OE-2010,Szwoch-IIMSS-2010,Tian-TSMC-2011}, while other approaches employ edge detection frameworks~\cite{Szwoch-MTA-2016,Dahi-CVIU-2017} or generative models~\cite{Wen-ICIP-2009}. Some frameworks model the abandoned objects detection problem as finite state automata~\cite{Kwak-OE-2010,Lin-ICPR-2014,Lin-TIFS-2015}, while others use temporal logic-based inference as an alternative solution~\cite{Bhargava-AVSS-2007,Bhargava-MVA-2009,Ferryman-PRL-2013}.
Different from all other works, Kong et al.~\cite{Kong-TIP-2010} try to detect abandoned objects on the road using a moving camera.
Ferryman et al.~\cite{Ferryman-PRL-2013} present a threat assessment algorithm that combines the concept of ownership with automatic understanding of social relations in order to infer abandonment of objects.
Similar to Porikli et al.~\cite{Porikli-JASP-2008}, Lin et al.~\cite{Lin-ICPR-2014,Lin-TIFS-2015} combine short-term and long-term background models to extract foreground objects.
Szwoch~\cite{Szwoch-MTA-2016} describes an algorithm for the detection of stable regions by comparing these regions with the contours of moving objects.
Dahi et al.~\cite{Dahi-CVIU-2017} present a method based on static edge detection and classification.
Pham et al.~\cite{Pham-ICISP-2017} propose a two-stage method for unattended object detection. The first stage tries to detect all possible abandoned objects, preventing false negatives. In the second stage, their method reduces false alarms by using similarity matching between first-stage candidates and the background model. Different from all previous methods, we use a cascade of convolutional neural networks in the second stage to recognize abandoned versus attended luggage items or other objects.
For a complete review of recent works on abandoned object detection, the reader is referred to the  survey of Cuevas et al.~\cite{Cuevas-CVIU-2016}.

\begin{figure}[!t]

\begin{center}
\includegraphics[width=0.94\linewidth]{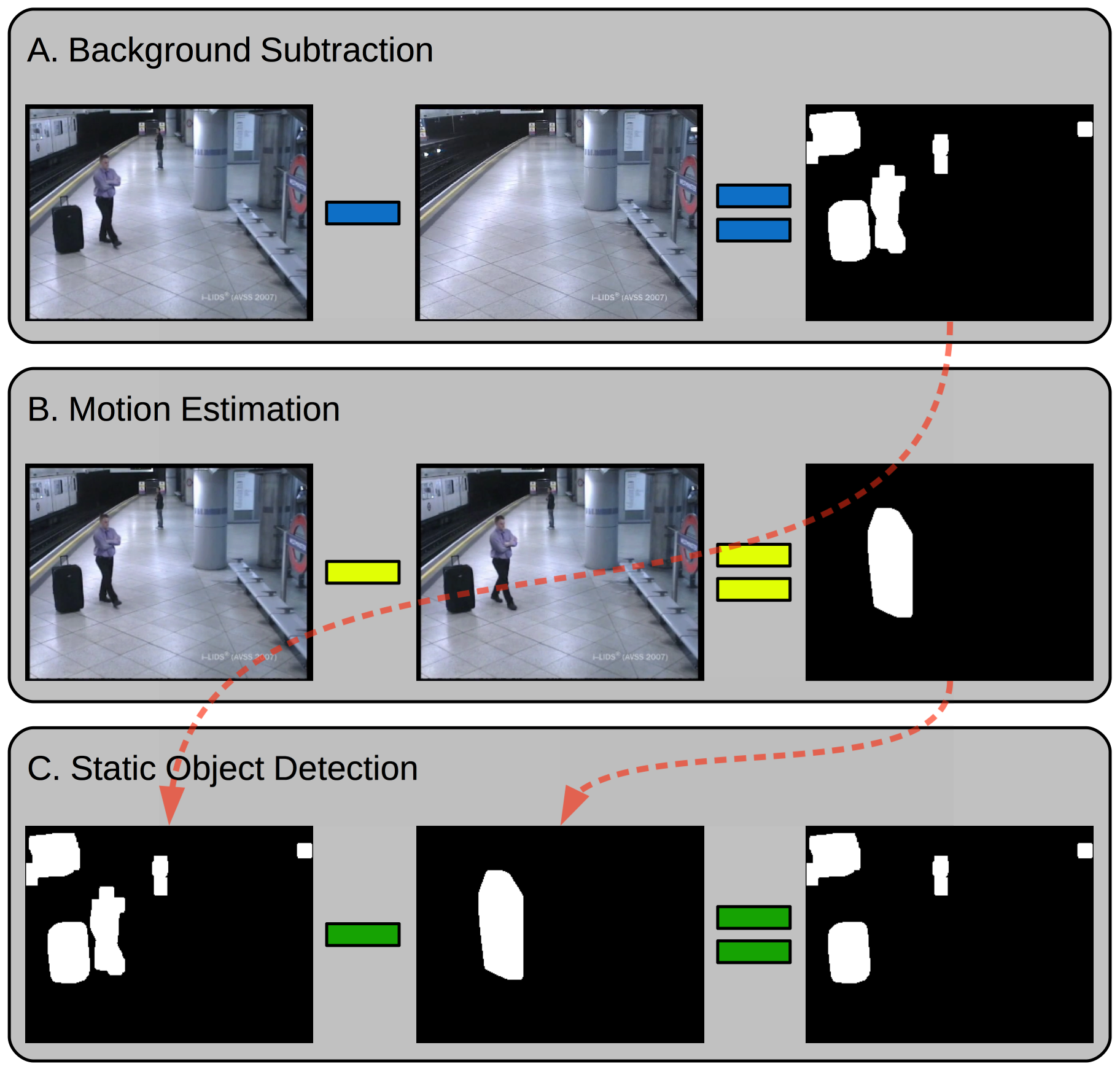}
\end{center}
\vspace*{-0.3cm}
\caption{Static object detection (SOD) pipeline used in the first stage of our approach for abandoned luggage detection. The pipeline is comprised of three steps. Step A: foreground estimation based on background subtraction. Step B: motion estimation based on subtracting temporally-close video frames. Step C: static object detection based on subtracting the motion mask from the foreground mask. In this particular example, the static objects are a subway train, a suitcase and two persons. Best viewed in color.}
\label{fig_first_stage_pipeline}
\vspace*{-0.2cm}
\end{figure}

\section{Method}
\label{sec_Method}
\vspace*{-0.2cm}

We propose a two-stage approach for abandoned luggage detection. In the first stage, we aim to detect all static objects in the scene. Among these objects however, there are many false positive detections, i.e. objects that do not represent abandoned luggage, e.g. persons standing still or even attended luggage items. Therefore, a second stage is necessary to distinguish between abandoned luggage items and other objects. We next describe the first stage in Section~\ref{sec_SOD} and the second stage in Section~\ref{sec_ALR}.

\vspace*{-0.1cm}
\subsection{Static Object Detection}
\label{sec_SOD}
\vspace*{-0.1cm}

Our \emph{static object detection} (SOD) approach is comprised of two components, namely background subtraction and motion estimation. The entire pipeline for the first stage is illustrated in Figure~\ref{fig_first_stage_pipeline}. The pipeline is based on three sequential steps: A, B and C. On each video frame, we first apply a standard method for background subtraction (step A) that simply subtracts the estimated background from each video frame. 
The resulted foreground mask is further processed by applying erosion and dilation filters. In the end, the foreground mask contains both static and moving objects. In order to find the moving objects in the scene, we estimate the motion (step B) by subtracting frames that are $5$ frames apart from each other. This will provide a mask representing the contour of moving objects. To fill the objects we apply erosion and dilation filters on the motion mask. We then apply a standard algorithm to find the connected components. For each connected component, we compute the convex hull. In the end, the motion mask contains the moving objects as convex connected components. To single out the static objects in the scene, we subtract the motion mask from the foreground mask and we obtain the static pixels mask (step C). On the static pixels mask, we compute the connected components. The resulted connected components represent the static objects in the scene. Finally, we determine the bounding box for each static object and extract the corresponding sub-image. The resulted images represent individual static objects. We track the static objects over multiple frames. 
We consider that two bounding boxes belong to the same track if the \emph{intersection over union} (IoU) is greater than $0.5$. Finally, the static object tracks are further processed in the second stage to determine if the objects are indeed abandoned luggage items.

\vspace*{-0.1cm}
\subsection{Abandoned Luggage Recognition}
\label{sec_ALR}
\vspace*{-0.1cm}

In the second stage, we employ a \emph{cascade of convolutional neural networks} (CCNN) to recognize abandoned luggage in the object tracks resulted from the first stage. We employ the GoogLeNet~\cite{Szegedy-CVPR-2015} architecture for both convolutional neural networks. We start from a GoogLeNet model that is pre-trained on the ILSVRC benchmark~\cite{Russakovsky2015}, and train only the last layer.

The first neural network is trained to recognize images containing luggage items, e.g. suitcases, hand bags, backpacks and so on. However, some of the test samples labeled as positive by the CNN may contain luggage items that are not abandoned. For instance, a person that stands still next to its own luggage (a common situation when one waits in a train station or airport) is likely to be detected as a static object in the first stage, and the neural network might activate if there is a luggage item in the corresponding image. Hence, this kind of situation is likely to be detected as an abandoned luggage item, until this point. Nonetheless, the second neural network in our cascade is specifically trained to solve such undesired situations. The network is given positive examples with abandoned luggage items and negative examples with attended luggage items, i.e. luggage items with people standing by. The second network is only applied on the image samples that are labeled as positive by the first network. In literature, this kind of pipeline is known as a \emph{cascade of classifiers}~\cite{Viola-2004}.

An important remark is that we need to extend the bounding box around the luggage item to detect if there are persons standing nearby with the second CNN. Let $h\times w$ be the size of a bounding box. For each image sample labeled as positive by the first CNN, we extend the bounding box to a size of $2h \times 3w$, and extract the corresponding larger image. To process the static object tracks in real-time, we apply the CCNN at every $10$ frames. The predicted classification scores for an object track are temporally smoothed using a Gaussian filter of $25$ frames ($1$ second). The sign transfer function is then applied to transform the scores for an object track into class labels. For each object track, we apply a majority voting scheme to determine the final class label for the respective object track.

\begin{figure}[!t]

\begin{center}
\includegraphics[width=0.9\linewidth]{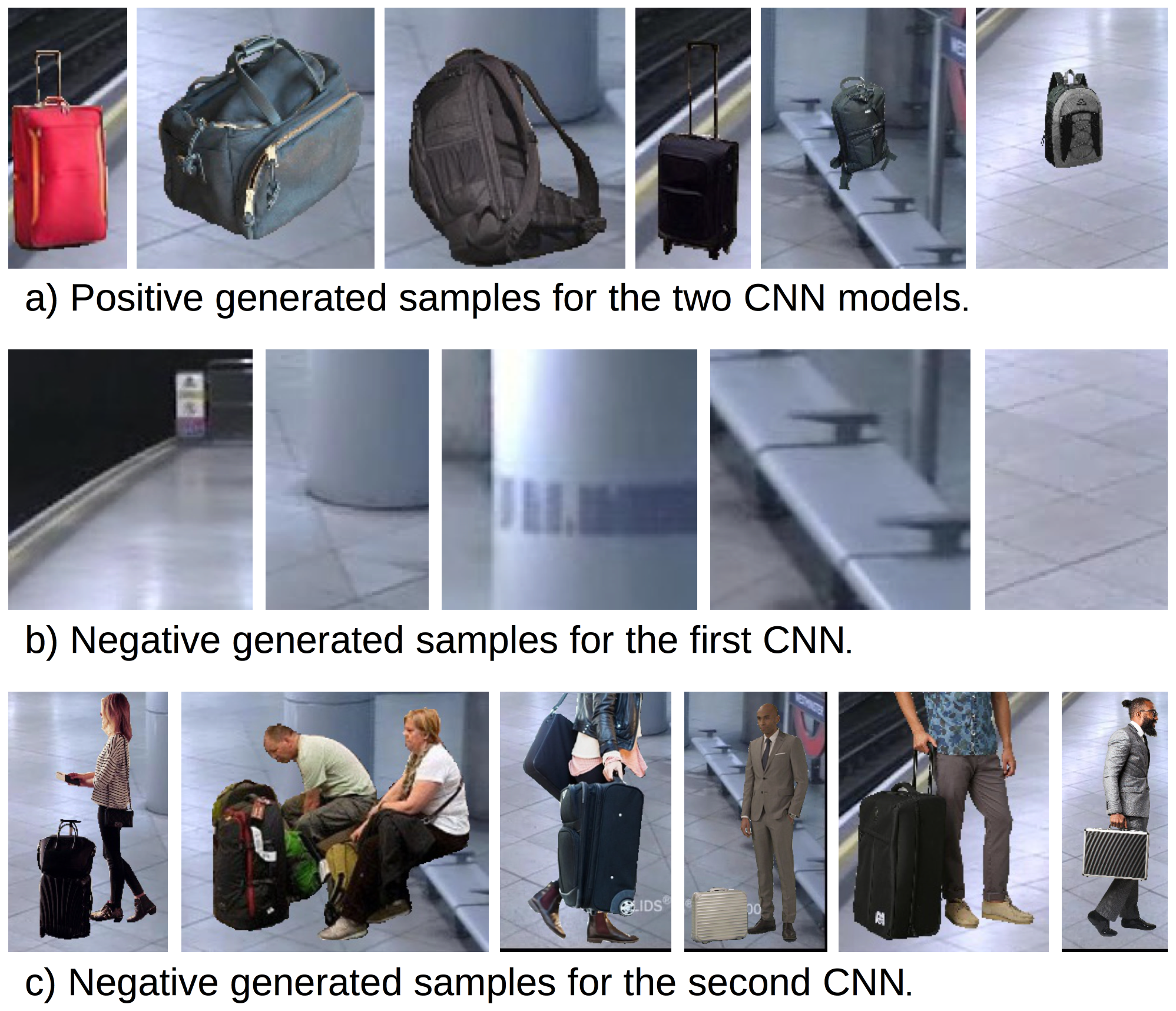}
\end{center}
\vspace*{-0.3cm}
\caption{Realistic image samples generated by randomly sampling sub-images from the estimated background image (second row) and by superimposing luggage items (first row) or people standing by their luggage items (third row). These samples are used to fine-tune our convolutional neural networks for a particular scene. Best viewed in color.}
\label{fig_generated_samples}
\vspace*{-0.2cm}
\end{figure}

To train both CNN models, we provide two types of examples. On one hand, we use images collected from the Internet in order to improve the generalization capacity of our neural networks. On the other hand, we want our models to be adapted to each individual scene in order to provide the best possible results in the respective scene. However, obtaining real image samples from each scene in order to fine-tune the networks is not a viable approach, as collecting these samples requires a large amount of time. For faster deployment, we propose to generate realistic image samples instead of collecting them. To fine-tune the first CNN model, we superimpose various template luggage items at random locations over the estimated background of each individual scene to obtain additional positive samples. We also select random sub-images from the background as negative examples. In a similar manner, we generate samples to train the second CNN model in our cascade. We use the same positive samples as for the first CNN model, but we generate the negative samples by superimposing people carrying or standing by their luggage items at various random locations over the estimated background. Figure~\ref{fig_generated_samples} illustrates a few image samples generated for one video scene considered in the experiments presented in Section~\ref{sec_Experiments}.

\begin{table}[!t]
\small{
\caption{Image classification results for our first CNN trained to distinguish abandoned luggage items from other objects, and our second CNN trained to distinguish abandoned luggage items from attended luggage items.}
\begin{center}
\begin{tabular}{|l|c|c|c|}
\hline
Method          & Precision	    & Recall	    & Accuracy\\
\hline
First CNN	    & $97.31\%$		& $82.12\%$		& $96.37\%$\\
Second CNN      & $96.96\%$		& $94.11\%$		& $95.36\%$\\
\hline
\end{tabular}
\end{center}
\label{tab_prelim_results}
}
\vspace*{-0.5cm}
\end{table}

\begin{table*}[!th]
\small{
\caption{Frame-level and pixel-level metrics for our approach based on static object detection (SOD) and a cascade of convolutional neural networks (CCNN) trained with and without generated samples versus a baseline approach based on CNN. The methods are evaluated on four data sets: AVSS 2007, PETS 2006, PETS 2007 and TCD (Trinity College of Dublin). The best score for each metric on each data set is highlighted in bold.}
\begin{center}
\begin{tabular}{|l|l|c|c|c|c|c|c|}
\hline
Data Set        & Method		                & \multicolumn{3}{|c|}{Frame-level}	                        & \multicolumn{3}{|c|}{Pixel-level}\\
\cline{3-8}
				&                                   & Precision	            & Recall		        & $F_1$ score	        & Precision		        & Recall		        & $F_1$ score\\
\hline
                & SOD + CNN	(baseline)	            & $60.17\%$		        & $60.87\%$		        & $60.52\%$		        & $41.19\%$		        & $53.03\%$		        & $46.37\%$\\
AVSS 2007       & SOD + CCNN		                & $\mathbf{97.77}\%$	& $51.89\%$		        & $67.80\%$		        & $\mathbf{97.77}\%$	& $51.89\%$		        & $67.80\%$\\
                & SOD + CCNN + Generated Samples    & $97.48\%$		        & $\mathbf{66.59}\%$    & $\mathbf{79.13}\%$    & $97.47\%$		        & $\mathbf{65.70}\%$	& $\mathbf{78.49}\%$\\
\hline
                & SOD + CNN	(baseline)              & $68.01\%$		        & $69.54\%$		        & $68.77\%$		        & $68.00\%$		        & $69.54\%$		        & $68.76\%$\\
PETS 2006       & SOD + CCNN		                & $83.25\%$		        & $69.54\%$		        & $75.78\%$		        & $83.25\%$		        & $69.54\%$		        & $75.78\%$\\
                & SOD + CCNN + Generated Samples    & $\mathbf{95.67}\%$	& $\mathbf{83.74}\%$	& $\mathbf{89.31}\%$	& $\mathbf{95.67}\%$	& $\mathbf{83.74}\%$	& $\mathbf{89.31}\%$\\
\hline
                & SOD + CNN	(baseline)              & $65.35\%$		        & $\mathbf{99.61}\%$	& $78.92\%$		        & $65.17\%$		        & $\mathbf{99.61}\%$	& $78.79\%$\\
PETS 2007       & SOD + CCNN		                & $69.13\%$		        & $\mathbf{99.61}\%$	& $81.62\%$		        & $68.99\%$		        & $\mathbf{99.61}\%$	& $81.52\%$\\
                & SOD + CCNN + Generated Samples    & $\mathbf{97.47}\%$	& $\mathbf{99.61}\%$	& $\mathbf{98.53}\%$	& $\mathbf{97.46}\%$	& $\mathbf{99.61}\%$	& $\mathbf{98.52}\%$\\
\hline
                & SOD + CNN	(baseline)              & $\mathbf{98.62}\%$	& $\mathbf{100}\%$		& $\mathbf{99.31}\%$	& $\mathbf{98.62}\%$	& $\mathbf{100}\%$		& $\mathbf{99.31}\%$\\
TCD             & SOD + CCNN		                & $\mathbf{98.62}\%$	& $\mathbf{100}\%$		& $\mathbf{99.31}\%$	& $\mathbf{98.62}\%$	& $\mathbf{100}\%$		& $\mathbf{99.31}\%$\\
                & SOD + CCNN + Generated Samples    & $\mathbf{98.62}\%$	& $\mathbf{100}\%$		& $\mathbf{99.31}\%$	& $\mathbf{98.62}\%$	& $\mathbf{100}\%$		& $\mathbf{99.31}\%$\\
\hline
                & SOD + CNN	(baseline)	            & $70.32\%$		        & $76.33\%$		        & $73.20\%$		        & $66.22\%$		        & $74.65\%$		        & $70.18\%$\\
Overall Average & SOD + CCNN		                & $86.54\%$		        & $74.41\%$		        & $80.02\%$		        & $86.52\%$		        & $74.40\%$		        & $80.00\%$\\
                & SOD + CCNN + Generated Samples    & $\mathbf{96.74}\%$	& $\mathbf{84.65}\%$	& $\mathbf{90.29}\%$	& $\mathbf{96.73}\%$	& $\mathbf{84.46}\%$	& $\mathbf{90.18}\%$\\
\hline
\end{tabular}
\end{center}
\label{tab_ALD_results}
}
\vspace*{-0.5cm}
\end{table*}

\section{Experiments}
\label{sec_Experiments}
\vspace*{-0.2cm}
\subsection{Data Sets}

We test our method on four datasets: AVSS 2007~\cite{i-LIDS-2006}, PETS 2006~\cite{PETS-2006}, PETS 2007~\cite{PETS-2007} and TCD~\cite{Dawson-Howe-2014}. In the i-LIDS bag and vehicle detection challenge (AVSS 2007) abandoned baggage data set, the detection areas are divided into near, middle and far, and there is one video per area. The PETS 2006 data set consists of seven scenarios captured from four different viewpoints (cameras). It contains multiple types of luggage: briefcase, suitcase, 25 litre rucksack, 70 litre backpack and sky gear carrier. We used the videos from a single viewpoint (third camera). 
In a similar fashion, we used the abandoned luggage scenarios (S7 and S8) from PETS 2007, using a single viewpoint (third camera). The Trinity College Dublin (TCD) data set contains two videos showing objects being abandoned and later collected. In total, there are $14$ test videos with a total of $42869$ frames.

\vspace*{-0.1cm}
\subsection{Evaluation}
\vspace*{-0.1cm}

We use three frame-level and pixel-level metrics for evaluation: precision, recall and $F_1$ score. \emph{Precision} is defined as the number of true-positive detections divided by the total number of detections, and \emph{recall} as the number of true-positive detections divided by the number of ground-truth instances. The $F_1$ measure (also known as the $F_1$ score) is the harmonic mean of precision and recall.
We define the frame-level and pixel-level metrics as in other works for abnormal event detection~\cite{Cong-CVPR-2011, Giorno-ECCV-2016,Mahadevan-CVPR-2010,Ionescu-ICCV-2017}. At the pixel-level, the corresponding luggage item is considered as being correctly detected if the IoU between the detected bounding box and the ground-truth bounding box is greater than $0.2$. At the frame-level, a frame is considered a correct detection if it contains at least one abandoned luggage item (there is no constraint regarding the overlap between the detected bounding box and the ground-truth bounding box). In order to calculate these metrics, we manually annotated all videos with ground-truth bounding boxes, since the data sets do not provide such annotations.

\vspace*{-0.1cm}
\subsection{Baseline and Models}
\vspace*{-0.1cm}

We consider as baseline a stripped-down version of our approach. The baseline is also based on two stages, with the first stage for static object detection identical to our own approach. In the second stage, we replace the cascade of convolutional neural networks with a single CNN. The baseline CNN is identical to the first CNN in our cascade. In the experiments, we consider two versions for the CCNN. The first version (SOD+CCNN) is trained using only the samples collected from the Internet. The second version (SOD+CCNN+Generated Samples) includes the generated samples in the training process. We present results with and without generated samples, to show the performance gain provided by the addition of generated samples.

\begin{figure}[!t]

\begin{center}
\includegraphics[width=0.75\linewidth]{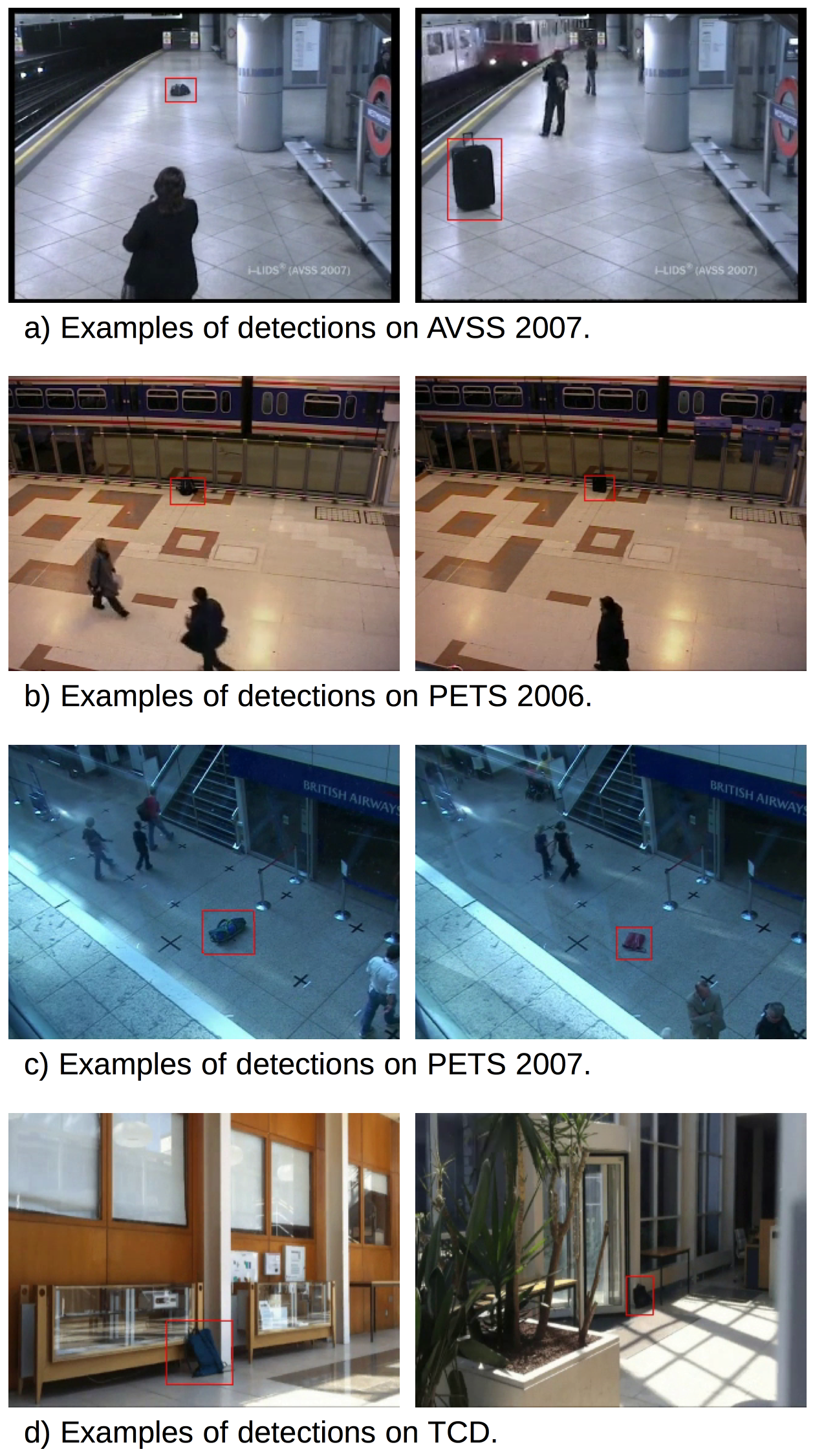}
\end{center}
\vspace*{-0.3cm}
\caption{Examples of abandoned luggage items detected by our approach based on SOD and CCNN trained with generated samples. Best viewed in color.}
\label{fig_detection_examples}
\vspace*{-0.2cm}
\end{figure}

\vspace*{-0.1cm}
\subsection{Results and Discussion}
\vspace*{-0.1cm}

\noindent
{\bf Preliminary classification results.}
We first train and test our convolutional neural networks on a collection of images collected from the Internet. In our data set, there are $2207$ images with abandoned luggage items, $2000$ with attended luggage items, and another $8035$ samples with other objects such as people, cars, buses, trains, and so on. The data set for the first CNN is formed of the images with abandoned luggage items as positive examples and the images with other objects as negative examples. The data set for the second CNN is formed of the images with abandoned luggage items as positive examples and the images with people attending luggage items as negative examples. Both data sets are split into $80\%$ for training and $20\%$ for testing. The training sets are augmented using flipped and blurred versions of the original images. 
The classification results presented in Table~\ref{tab_prelim_results} indicate that both methods attain precision levels of around $97\%$ and accuracy rates higher than $95\%$. The first CNN in our cascade attains a lower recall of almost $82\%$. Overall, the classification results indicate that both networks are well trained and ready to be used in practice.

\noindent
{\bf Results for abandoned luggage detection in video.}
Table~\ref{tab_ALD_results} shows the results of our approach trained with and without generated samples against a strong CNN baseline. On the AVSS 2007 data set, both CCNN approaches yield considerably better frame-level and pixel-level precisions compared to the baseline CNN. The best $F_1$ scores ($79.13\%$ at the frame-level and $78.49\%$ at the pixel-level) on AVSS 2007 are obtained by the CCNN version that is trained with generated samples. The same CCNN version also attains the best $F_1$ scores on PETS 2006, with an improvement higher than $20\%$ over the baseline. On PETS 2007, all methods obtain the same recall scores (above $99\%$). However, the CCNN version trained with generated samples reaches much better precision levels ($97.47\%$ at the frame-level and $97.46\%$ at the pixel-level) than the baseline CNN. The TCD data set seems to be quite easy, since all the evaluated methods obtain scores higher than $98\%$ for all metrics. Table~\ref{tab_ALD_results} also includes the average for each metric computed on the entire $14$ videos from all four data sets. We can observe that on average, the CCNN approach trained without generated samples yields average $F_1$ scores that are almost $10\%$ higher than the baseline CNN. However, the CCNN approach trained with generated samples attains even higher average $F_1$ scores. With an $F_1$ score of $90.29\%$ at the frame-level and an $F_1$ score of $90.18\%$ at the pixel-level, our best approach is roughly $20\%$ better than the baseline. We note that our framework runs at nearly $40$ frames per second on a machine with Intel Xeon Processor E5 $1.7$ GHz CPU and $32$ GB of RAM, without using parallel threading. 

\vspace*{-0.1cm}
\section{Conclusion}
\label{sec_Conclusion}
\vspace*{-0.2cm}

In this paper, we have presented a novel approach for abandoned luggage detection in video that works in real-time. We compared our approach with a strong CNN baseline. The empirical results indicate that employing a cascade of two CNN models trained on collected as well as generated image samples provides improvements above $8\%$ for all evaluation metrics.

\ifCLASSOPTIONcompsoc
\vspace*{-0.1cm}
\section*{Acknowledgments}
\vspace*{-0.1cm}
The work of Radu Tudor Ionescu is supported through project grant PN-III-P2-2.1-PED-2016-1842.
The work of Sorina Smeureanu is supported by SecurifAI through Project P/38/185 funded under POC-A1-A1.1.1-C-2015.
\else
\vspace*{-0.1cm}
\section*{Acknowledgment}
\vspace*{-0.1cm}
The work of Radu Tudor Ionescu is supported through project grant PN-III-P2-2.1-PED-2016-1842.
The work of Sorina Smeureanu is supported by SecurifAI through Project P/38/185 funded under POC-A1-A1.1.1-C-2015.
\fi



\bibliographystyle{IEEEtran}
\bibliography{IEEEabrv,references}

\end{document}